\documentclass[letterpaper, 10 pt, conference]{ieeeconf}  % Comment this line out if you need a4paper

\IEEEoverridecommandlockouts                              % This command is only needed if 
\usepackage{graphics} % for pdf, bitmapped graphics files
\usepackage{epsfig} % for postscript graphics files
\usepackage{mathptmx} % assumes new font selection scheme installed
\usepackage{times} % assumes new font selection scheme installed
\usepackage{amsmath} % assumes amsmath package installed
\usepackage{amssymb}  % assumes amsmath package installed
\usepackage{subfigure}
\usepackage[utf8]{inputenc}
\usepackage{dirtytalk}
\usepackage{color,soul}
\usepackage{algorithm}  
\usepackage{algpseudocode}  
\usepackage{amsmath}  
\usepackage{subfig}
\usepackage{cite}
\usepackage{multirow}
\usepackage{verbatim}
\usepackage{subfigure}
\usepackage{diagbox}
\usepackage{caption}
\usepackage{tabularx}
\usepackage{makeidx}
\usepackage{booktabs}
\makeatletter

\newcommand{\Rmnum}[1]{\expandafter\@slowromancap\romannumeral #1@}
\makeatother
\title{\LARGE \bf
Robot Navigation with Reinforcement Learned Path Generation and Fine-Tuned Motion Control
}

\author{Longyuan Zhang$^{1}$, Ziyue Hou$^{1}$, Ji Wang$^{1}$, Ziang Liu$^{1}$ and Wei Li$^{1}$
\thanks{$^{1}$Longyuan Zhang, Ziyue Hou, Ji Wang, Ziang Liu and Wei Li are with Academy for Engineering and Technology,
        Fudan University, Shanghai, China.
        }%
}%

\begin{document}

\maketitle
\thispagestyle{empty}
\pagestyle{empty}

%%%%%%%%%%%%%%%%%%%%%%%%%%%%%%%%%%%%%%%%%%%%%%%%%%%%%%%%%%%%%%%%%%%%%%%%%%%%%%%%
\begin{abstract}

In this paper, we propose a novel reinforcement learning (RL) based path generation (RL-PG) approach for mobile robot navigation without a prior exploration of an unknown environment. Multiple predictive path points are dynamically generated by a deep Markov model optimized using RL approach for robot to track. To ensure the safety when tracking the predictive points, the robot's motion is fine-tuned by a motion fine-tuning module. Such an approach, using the deep Markov model with RL algorithm for planning, focuses on the relationship between adjacent path points. We analyze the benefits that our proposed approach are more effective and are with higher success rate than RL-Based approach DWA-RL\cite{DWARL} and a traditional navigation approach APF\cite{APF}. We deploy our model on both simulation and physical platforms and demonstrate our model performs robot navigation effectively and safely.
\end{abstract}
% motion fine-tuning module

%% fine-tune module mention 

%%%%%%%%%%%%%%%%%%%%%%%%%%%%%%%%%%%%%%%%%%%%%%%%%%%%%%%%%%%%%%%%%%%%%%%%%%%%%%%%
\section{Introduction}
The path planning task is one of the most common tasks for mobile robots or autonomous vehicles. Especially in an unknown environment without mapping in advance, how a robot manages collision avoidance, trajectory smoothing, and avoiding sub-optimal solutions are challenging during path planning.

When faced with a completely unfamiliar environment, the implementation of most traditional map-based approaches such as A* and RRT algorithms would become difficult. Thus some approaches in which the map information is not required have been proposed to solve path planning challenges (e.g., \cite{DWA, MPC}). 
However, several difficulties still prevent traditional algorithms from obtaining a breakthrough, since the traditional approaches suffer from slow computational speed when faced with complex environmental situations. Traditional algorithms rely on mathematical computation and inference, which may cost more time and greatly affect the speed of planning. Another defect is that the traditional path planning algorithms are prone to be affected by noise-filled raw sensor data, which will lead to a more difficult deployment of traditional path planning approaches on physical robots.

To solve the shortcomings of traditional algorithms in path planning problems, many learning-based approaches (e.g., \cite{IJRR, PRIMAL, DWARL}) have been proposed. Some imitation learning-based methods (e.g., \cite{carrace}) have achieved fast inference speed in drone, autonomous driving, and mobile robot navigation tasks. However, it requires a large amount of training data. The expert trajectories in data sets are not guaranteed to be the most optimal. More importantly, the student network may not learn the corresponding trajectories with unseen observation spaces, which may lead to unexpected behavior. Some other approaches utilize the RL-based End-to-End approaches (e.g., \cite{IJRR, DWARL, PRMRL, PRMRL2}) to learn from partial observation space to directly output driving commands. However, an important issue is that RL-based End-to-End approaches are hard to generalize, since once the model is trained, it is hard to correct the robot's motion commands when collisions happen.
% e2e cannot adjust

\begin{figure*}[t]
       \centering
       \includegraphics[width=0.9\textwidth]{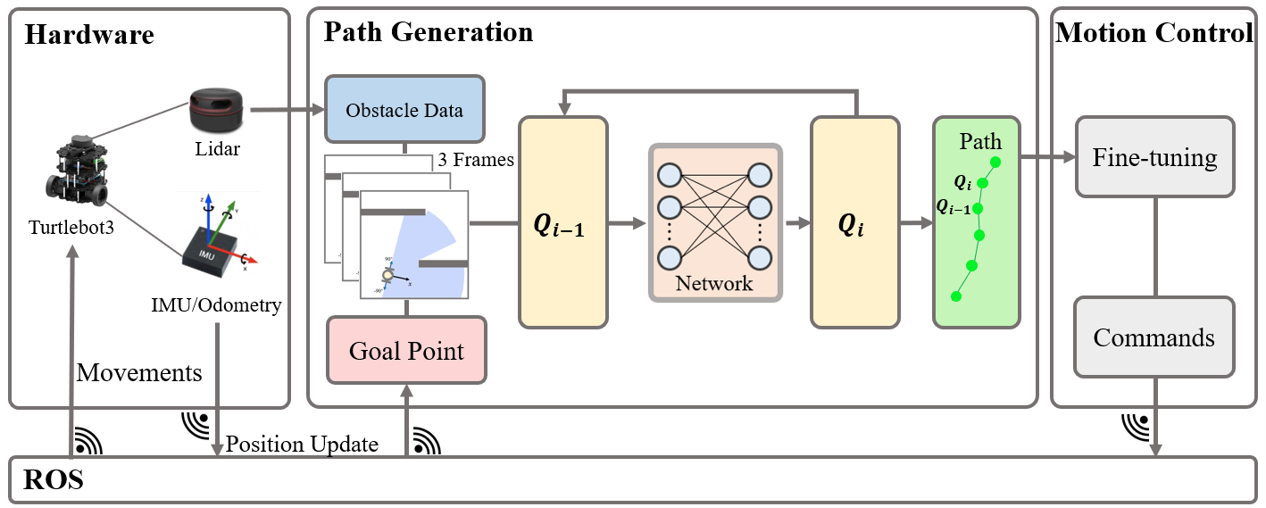}
       \caption{An overview of the framework. The obstacle data is collected by the lidar on Turtlebot3. The goal point is set up in advance. $Q_{i-1}$ and $Q_i$ represent the two adjacent path points. The network is a deep Markov model based policy network. During the robot movement, the path generation module dynamically generates multiple local paths based on varying sensor information and positions. The motion control module fine-tunes the paths and then sends execution commands to the robot.}
       \label{fig:overview}
\end{figure*}
Here we present a RL-based predictive path generation approach with fine-tuned motion control to drive the mobile robots in a variety of complex environments without any prior exploration while having only access to onboard sensors and computation. Different from other RL-based End-to-End approaches to output motion commands, we propose a novel RL-based method to generate predictive path points. This formulation could firstly fully utilize the advantages of learning-based methods such as fast inference speed and robustness to learn the mapping from raw sensor data to various types of outputs. Once a model is trained, substituting its controller in a modular way could fulfill navigation task requirements. Another great benefit is that our approach manages to decouple the trajectory generation and motion control since our RL-based approach's action space and credit assignment are both based on planned trajectory, which means our path generation method is more robust to various environments. 

Our novel actor policy based on the deep Markov model is designed for predictive path point generation. When generating paths, we fully consider the sequential relationship of adjacent path points. During the path generating process, the robot’s position, posture, and sensor information are dynamically changed. 
Due to robot kinematic limitations, the robot's motion may not be consistent with the generated path, resulting in collision.
Thus the motion fine-tuning module is proposed to fix the problem and improve safety. The overall framework is shown in Figure \ref{fig:overview}.

The main contributions of our work include:
\begin{itemize}
    \item
     A RL-based path generation with fine-tuned motion control is proposed for robot's navigation in an unknown environment without prior exploration. The path points prediction enables navigation and collision avoidance, while motion control is only used to ensure the safety of the robot in case of emergency.
    \item A novel deep Markov model trained by deep reinforcement learning to dynamically and iteratively predict planning path points. That is, at each time step, each predictive path point is obtained based solely on the previous path point and partially observable space, and we treat this point set as a planning trajectory. This predictive reinforcement learning based `action' space later in simulation and experiments proves to be more optimal when the robot explores and navigates in unknown environments.
    %% so detailed. 
    \item Combining predictive path generation with control has the advantage of low coupling, allowing better differentiation between the two functions, and the role of each module on the robot will be well played. 
    
\end{itemize}
\section{Related Work}
\begin{figure}[t]
       \centering
       \includegraphics[width=0.4\textwidth]{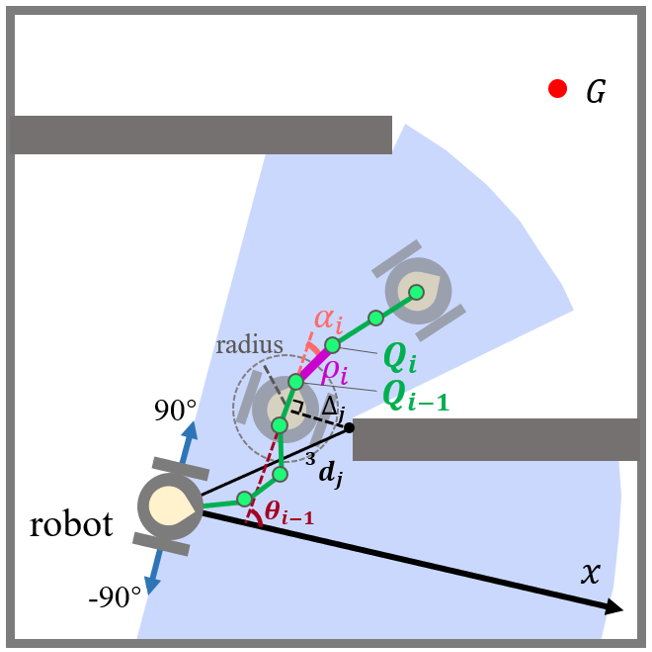}
       \caption{The blue sector is the obstacle data and the green points are the path points generated by the robot through the reinforcement learning method. $G$ represents the goal point. A plane rectangular coordinate system is established with the robot itself as the origin and x-axis denotes the robot's facing direction. $\rho_i, \alpha_i$ are the parameters of $Q_i$, which represents the distance the robot needs to travel and the angle that needs to rotate based on the $(i-1)^{th}$ path point $Q_{i-1}$. The $^3d_j$, defined in Equation \ref{eq:ORD} of Section IV, represents one of the obstacle data (distance between scanned obstacle and the lidar), while $\Delta_j$ represents the corresponding distance between the obstacle and the path.} 
       \label{fig:radar_and_points}
\end{figure}
% Our work covers many aspects including RL-bas
% \subsection{Path Generation}

For the representation of paths, the traditional approaches (e.g., \cite{ego}) are mostly points and line segments. Splitting the path into points and line segments is convenient for path expression and calculation. 
Traditional methods of path planning (e.g., \cite{onlineMPC, DWA, traplanning}), typically generate a predicted segment of trajectory based on its state and convert it into executable instructions. 
As the number of predicted nodes increases, the computation time and resource consumption of the whole algorithm increase dramatically.

A number of path planning methods combined with learning algorithms require prior exploration of environments\cite{priorexplo}. For example, the long-range path planning method (PRM-RL) \cite{PRMRL, PRMRL2} uses a traditional method PRM for path planning of a globally known map, and then uses reinforcement learning method to generate robot's actions for movement. RL-RRT\cite{RL-RRT}, similar to PRM-RL, uses RRT algorithm to plan the global path and uses RL method to control robot's motion for dynamic obstacle avoidance. Other grid-map-based methods such as PRIMAL\cite{PRIMAL}, Improved DynaQ\cite{dyna-q}, RL-basd Heuristic Search\cite{HeuristicSearch}, UAV-Path-Planning\cite{UAVpp}, etc., search for the higher scoring grids using learning algorithms in the grid map to form a path. Another approach uses Q-Learning\cite{RLANN} in the ANN method to plan the robot's global moving direction, but switches to robot local planing module when it encounters an obstacle and guides the robot to change its direction to avoid collision.

Some other similar studies focus on using RL based method to generate the global path. In the known map, studies (e.g., \cite{waypointplanning, RLGLOBAL, hierarchical}) present approaches based on RL that train networks with global information to generate all path points from the start point to the goal point once for the robots to follow. 

Therefore, such RL-based approaches mentioned above focus on a prior exploration to the environment, planning the path and using reinforcement learning methods to regulate the robots' movement actions. Instead, our approach focuses on utilizing reinforcement learning methods to make local path planning using obstacles information scanned by robot's lidar without a map built in advance, and use fine-tuned motion control for robot movement.

Some other methods that do not require a prior exploration of the map, such as a local planner trained by RL\cite{IJRR, DWARL}, only plan the robot's motion instructions. PointGoal navigation, proposed in DDPPO\cite{DDPPO}, is used to infer the robot's forward or rotation actions based on the images and GPS information using PPO method by giving robot the relative position of the target point. Here, the global map is of low importance and mainly assists the robot to locate the goal position. 
% Many approaches are available for navigation based on reinforcement learning. The study of quadrotors \cite{quadrotors} solves the path following problem by training different controllers using RL. Some studies (e.g., \cite{MPCRLtrajopti, IDC}) use traditional algorithms to generate a path, and then use reinforcement learning to optimize the path structure so that the path becomes smoother or more optimal than the path generated by traditional algorithms. The focus of their works are path optimization rather than path generation. Some studies (e.g., \cite{globalRL, IJRR}) use the End-to-End approach based on RL to accomplish obstacle avoidance or pathfinding tasks. These approaches, however, focus on optimizing existing paths or robot motions instead of path generation.

% The framework combining path generation and control is widely used. In the car racing scene \cite{carrace}, an vision-based approach combining imitation and reinforcement learning methods for autonomous car race is proposed. In the crossroads scenario\cite{IDC}, a path is generated and the controller is trained using RL approach to avoid collision. They improve the efficiency of robot path planning by subdividing functional modules. Such a framework can reduce the impact of the computational speed of the planning algorithm, improve the efficiency of the system, and ensure robot safety to a greater extent.
% For motion control, different controllers (e.g., \cite{APF, APF1, PID}) are adapted to track the generated paths.

\section{Problem Formulation}
% \subsection{Overall Framework}
% We elaborate on the framework and specific details of our approach. The overview of the framework is shown in Figure \ref{fig:overview}. The framework is divided into two main parts, including the path generation module and the motion control module.

% In the planning part, we use reinforcement learning approach combined with deep Markov model to generate paths. Our model uses the environmental information and the previous output $Q_{i-1}$ as the input to generate $Q_i$, which is a dynamic planning and recycling process.

% In the control part, a reference path data generated by RL is transmitted to the controller. The controller activates asynchronously after a certain time interval, instead of changing the reference path of the controller for each generated path. At this point, the controller combines the environmental potential field and the reference path to take control and make corrections to the robot's motion.

% \subsection{Path Definition}

We here define how the robot's path is expressed and transformed in the coordinate frames used in our approach. We use both an absolute world coordinate frame and a robot-relative local polar coordinate in our definition.
% We here define how the path is expressed and how the path is transformed in the coordinate. We use two coordinate systems (the world coordinate system and the individual robot local polar coordinate system) in our definition. 

We use a set of points to represent a path. In the world coordinate system, the position and orientation of the robot is defined as $Qw(X, Y, \theta)$, where $X$, $Y$ represent the coordinates of the robot in the world coordinate system and $\theta$ represents the orientation of the robot. When converting $Qw$ into robot's local polar coordinate $Q_0$, its local polar coordinates can be expressed as $Q_0(0, 0)$. We define the robot's facing direction as the direction of the x-axis in the robot's local polar coordinate system. The orientation $\theta_0$ in polar coordinate at $Q_0$ is 0.

Then we define a generated path in the local coordinate as a set
\begin{equation}
    L = \{ Q_i(\rho_i, \alpha_i)|i\in(1,2,\dots,n)\}.
    \label{equation:the_path}
\end{equation}
Here $\rho_i$ represents the displacement of $Q_i$ with respect to the previous point $Q_{i-1}$, $\alpha_i$ represents the angular deflection of $Q_i$ with respect to the previous point $Q_{i-1}$, shown in Figure \ref{fig:radar_and_points}.

In another word, $Q_i(\rho_i, \alpha_i)$ represents the predicted distance the robot needs to travel and the predicted angle the robot needs to rotate based on the $(i-1)^{th}$ path point $Q_{i-1}(\rho_{i-1}, \alpha_{i-1})$ and the orientation $\theta_{i-1}$, shown in Figure \ref{fig:radar_and_points}. Here the orientation  $\theta_i$ is accumulated by all $\alpha$. 
\begin{equation}
\theta_i = \sum_{j=1}^i (\alpha_j).
\end{equation}

We calculate the world coordinates of point $i$ using $\rho_i$, $\theta_i$ and its previous predicted point state $Qw_i(x_{i-1}, y_{i-1}, \theta w_{i-1})$. Here $x_{i-1}$ and $y_{i-1}$ are the world coordinates of previous points $Qw_{i-1}$, and $\theta w_{i-1}$ is accumulated orientation of all the previous points in the world coordinates. Specially, for the first point $Qw_1$, $x_0$, $y_0$ and $\theta w_0$ represent robot's initial state.

When $i\geq1$, for the $i^{th}$ point, we have
\begin{equation}
\begin{split}
{x_i}={x_{i - 1}} + {\rho _i}\cos ({\theta w_i}),\\
{y_i}={y_{i - 1}} + {\rho _i}\sin ({\theta w_i}).
\end{split}
\end{equation}
So in the world coordinate we have
\begin{equation}
    Lw = \{Qw_i(x_i,y_i,\theta w_i)|i\in(1,2,\dots,n)\}.
\end{equation}
We use the robot's local coordinate system to generate the robot's path point set $Q$ corresponding to itself and then transform the point set in the world coordinate for the controller to track. 

For the goal point, we define it as $G(\rho_g, \theta_g)$ in the robot's local polar coordinate. For the obstacle data, we use 180-dimensional obstacle data in 3 consecutive frames scanned by lidar. %We use the latest 3 frames of obstacle data as another input and it will not change until the lidar has finished a full turn.

\section{Approach}

\subsection{Policy Representation}
\subsubsection{Observation Space}
%In the robot's local coordinate system, we use polar coordinates to generate points. 
For each path point $Q_i$, the observation $\textbf{o}^t_i$ consists of three parts: the latest 3 frames of obstacle data $\textbf{o}^t_d$, the goal point $\textbf{o}^t_g$, and the previous point $Q_{i-1}$. Obstacle data is obtained from a rotating lidar sensor which returns distance to obstacles. Lidar is sampled every 1 degree from -90 degree to 89 degree with 0 degree as the robot's forward heading. Three complete rotations of the lidar can then be combined into a history of three obstacle data readings as the obstacle data matrix $\textbf{o}^t_d$. 
% For each path point $Q_i$, the observation $\textbf{o}_i^t$ consists of three parts: the latest 3 frames of obstacle data $\textbf{o}_d^t$, the goal point $\textbf{o}_g^t$, and the previous point $Q_{i-1}$. In the sight of the robot, the radar information is obtained by the interaction of the radar sensor with the surrounding environment. As the position of the robot changes and the lidar completes a turn, the radar information can be completely updated once. When the lidar completes three turns, we can obtain three chronological dynamic radar information. Take the robot facing direction as 0° and sample the data every 1°. The data sampled every 1° is the distance of the lidar to the nearest obstacle at the corresponding angle. Take the three chronological radar information data from the lidar (-90°, 89°) as obstacle data matrix $\textbf{o}_d^t$.
\begin{equation}
\textbf{o}_d^t = 
    \begin{bmatrix}
    ^1D\\
    ^2D\\
    ^3D
    \end{bmatrix}, 
    ^nD = \{^nd_{i}|i\in [-90, 90)\cap \mathbb{Z}, n=1,2,3\}.
    \label{eq:ORD}
\end{equation}
Here $i$ represents the $i^{th}$ angle divided equally by 180-degree from -90° to 89°, $n$ represents the $n^{th}$ frame of the obstacle data, and $^nd_i$ represents the distance between the lidar and the obstacle scanned on the angle $i$ in the $n^{th}$ frame. As the robot is moving, $^nD (n=1,2,3)$ are scanned at different positions and postures.

The goal point information needs to be transformed. We convert the goal point from the world coordinate to the local polar coordinate according to the goal point position $Gw(x_{gw}, y_{gw})$ calibrated in the world coordinate system. We define the current state information of the robot mentioned in Section III as $Qw(X,Y,\theta)$. Here we have
\begin{align}
    &x_g = (x_{gw}-X)\cos{\theta} + (y_{gw}-Y)\sin{\theta}.\\
    &y_g = -(x_{gw}-X)\sin{\theta} + (y_{gw}-Y)\cos{\theta}.\\
    &(\rho_g, \theta_g) = (\sqrt{x_g^2+y_g^2}, \arctan(y_g/x_g)).
\end{align}
Thus we have the goal information in local polar coordinate $\textbf{o}_g^t(\rho_g, \theta_g)$.

Therefore, the observation space $\textbf{o}^t$ consists of three parts.
\begin{equation}
    \textbf{o}_i^t = (\textbf{o}_d^t, \textbf{o}_g^t, Q_{i-1}).
    \label{eq:obs}
\end{equation}
\subsubsection{Action Space}
In each cycle when generating a path, the path point $Q_i$ is one of the actions according to the data in the observation space and the network. Here the action space $\textbf{a}^t$ can be represented by $Q_i$.
During the process of generating the path, all the actions are combined to be a complete path. According to the definition of the path in Equation \ref{equation:the_path}, we have
\begin{equation}
    \textbf{a}^t = \{Q_i|i\in(1,2,\dots,n)\}.
    \label{eq:acts}
\end{equation}
\subsubsection{Reward Structure}
We design the reward by judging the three elements of the path point. We call these three parts as
\begin{itemize}
    \item $r_{c}$, if the predicted path collides with obstacles
    \item $r_{n}$, if the robot approaches the goal point
    \item $r_{s}$, smoothness judgment
\end{itemize}
% \begin{figure}[t]
%       \centering
%       \includegraphics[width=0.30\textwidth]{figures/points_radar_distance.png}
%       \caption{$d^{t3}_{\theta_{r}}$ is the ${\theta_r}^{th}$ data value of the third frame in $RD$. $T(\rho_t, \theta_t)$ represents the goal point in local coordinate. $Dpr$ represents the intersection of the ${\theta_r}^{th}$ piece of data with the obstacle. $\Delta_{pr}$ is the distance between the path and the ${\theta_r}^{th}$ scanned point $Dpr$. If the obstacles are out of the radar range, the $Dpr$ is regarded to be infinity. The conflicts will happen when any of 180 $Dpr$s are less than the radius of the robot.}
%       \label{fig:radar_distance}
% \end{figure}
The first is whether the generated path collides with the obstacle scanned by the lidar. We need to judge whether the path generated by the robot conflicts with the obstacles, as shown in Figure \ref{fig:radar_and_points}. For each frame of data swept by the lidar, we compute the scanned point and compute the distance $\Delta$ between the scanned point and the line segment connected by the path point. If $\Delta$ is less than the robot's radius, it is judged as a collision. 
\begin{equation}
r_{c} = 
\begin{cases}
-15, \;break \quad &if \; \Delta_i<radius \; , \; i\in [-90,90)\cap \mathbb{N}\\
0, \quad &otherwise.
\end{cases}
\end{equation}
If a collision occurs, we set $r_c=-15$, and terminate the training of the current process, which determines that the task failed. 

For whether the path is approaching the goal, we compare the distance from the path points to the goal point and the distance from the robot to the goal point.
For each point $i$, $s_i$ represents the distance between the $i^{th}$ path point and the goal point, and $d$ represents the distance between the robot and the goal point. If $s_i$ is smaller, the path point is closer to the goal, and the feedback is positive. Otherwise, it is negative. Thus
\begin{equation}
r_{n} = \sum_{i=1}^n (\frac{d-s_i}{i}).
\end{equation}
For the smoothness judgment, if the second parameter $\alpha_i$ of the generated path points $Q_i(\rho_i, \alpha_i)$ is large, it means that the angle the robot needs to turn is large. Then, we can limit the size of the angle of each path point $\alpha_i$ to solve the problem of path smoothness. Thus, for all points in the path, 
\begin{equation}
r_{s}=-\lambda\sum_{i=1}^n\alpha_i^2, \lambda = 0.0005.
\end{equation}
Combining $r_c$, $r_n$ and $r_s$, we will obtain the total reward
\begin{equation}
r = r_{c} + r_{n} + r_{s}.
\end{equation}
\subsubsection{Actor-Critic Network}
%% introduce PPO first
%% input output 
%% probability.

Our policy network is trained and inferenced in an iterable fashion. Given the input $\textbf{o}^t$ mentioned in Equation \ref{eq:obs} and output $\textbf{a}^t$ mentioned in Equation \ref{eq:acts}, our policy could iteratively compute the mapping from observation space $\textbf{o}^t$ to action space $\textbf{a}^t$.

Note that not all observation space is observable. After the first iteration, path points are no longer obtainable. Only at the first iteration, our $Q_{i}$ as part of observation space is obtained directly as the point that represents the current robot location. For the rest of the iteration steps, we mark those unobservable positional points as `virtual' position states that are generated by the previous iteration step, meaning the current single trajectory point depends on and only on partially observable environment space (in our space, lidar scan) and assume positional status at current iteration step which is exactly the output from the previous step. If $\pi$ represents the policy, we can generate all $Q$s from
% \begin{align}
%     &Q_1 = \pi(o_d^t, o_g^t, Q_0)\\
%     &Q_2 = \pi(o_d^t, o_g^t, Q_1) = \pi(o_d^t, o_g^t, \pi(o_d^t, o_g^t, Q_0)
% \end{align}
% $\dots$
\begin{align*}
    Q_i &\sim \pi(Q_i | (\textbf{o}_d^t, \textbf{o}_g^t, Q_{i-1})) \\ 
    &\sim \pi(Q_i | (\textbf{o}_d^t, \textbf{o}_g^t, \pi(Q_{i-1} | (\textbf{o}_d^t, \textbf{o}_g^t, Q_{i-2}))))\\
    &\sim \pi(Q_i | (\textbf{o}_d^t, \textbf{o}_g^t, \pi(\dots\pi(Q_1 | (\textbf{o}_d^t, \textbf{o}_g^t, Q_0))))). \tag{15}
\end{align*}
\begin{figure}[t]
       \centering
       \includegraphics[width=0.46\textwidth]{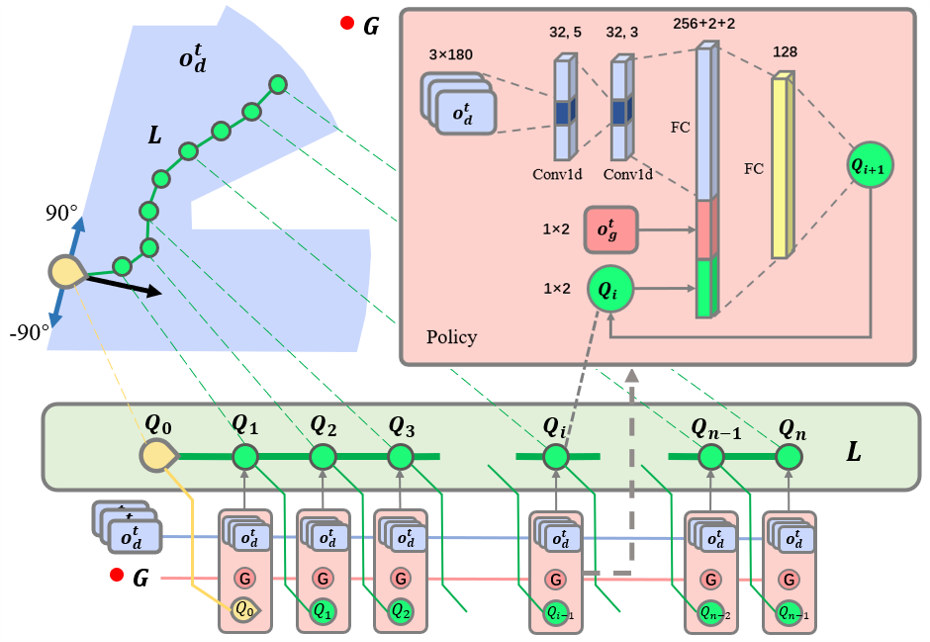}
       \caption{The architecture of the deep Markov model based policy network when generating path $L$. In the policy network, the inputs are the 3 frames of obstacle data $\textbf{o}_d^t$, local goal point $T$, and the previous point $Q_{i-1}$. It outputs current point $Q_i$. After $n$ times of point generation process, a complete path point set is presented. This process is then executed in a loop until the generation of $n$ path points is completed.}
       \label{fig:RNN}
\end{figure}
\begin{figure}[t]
       \centering
       \includegraphics[width=0.48\textwidth]{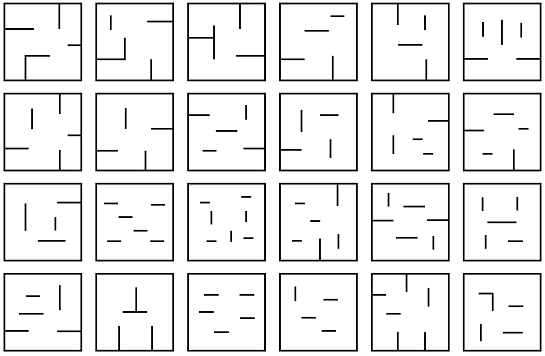}
       \caption{Training maps for 24 processes. The start points and the goal points are set randomly.}
       \label{fig:trainingmap}
\end{figure}
\begin{figure}[htbp]
       \centering
       \subfigure[]{\includegraphics[width=0.08\textwidth]{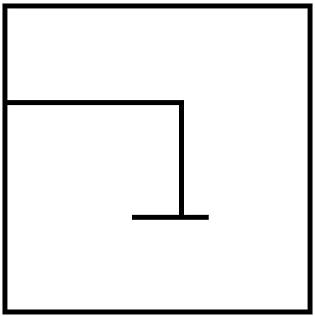}}
       \subfigure[]{\includegraphics[width=0.08\textwidth]{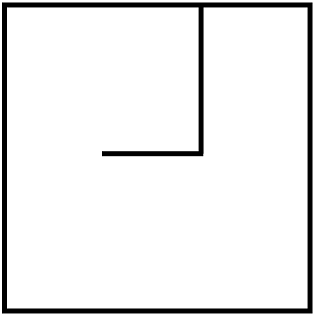}}
       \subfigure[]{\includegraphics[width=0.08\textwidth]{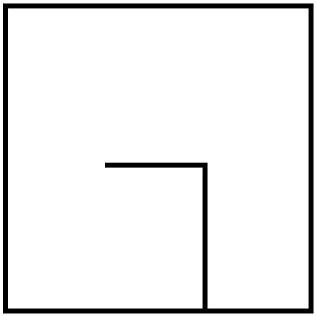}}
       \subfigure[]{\includegraphics[width=0.08\textwidth]{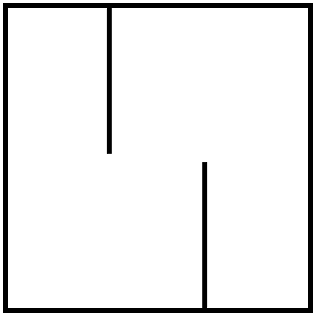}}
       \subfigure[]{\includegraphics[width=0.08\textwidth]{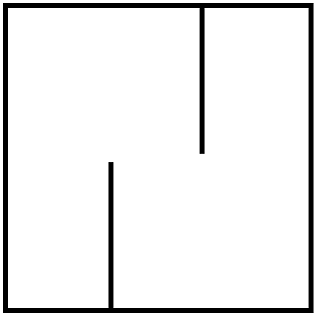}}
       \subfigure[]{\includegraphics[width=0.08\textwidth]{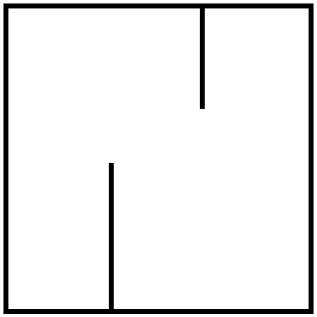}}
       \subfigure[]{\includegraphics[width=0.08\textwidth]{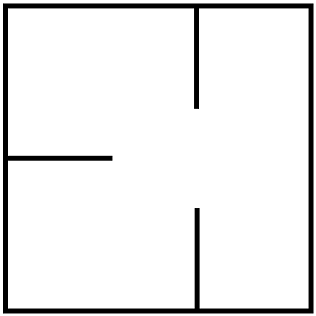}}
       \subfigure[]{\includegraphics[width=0.08\textwidth]{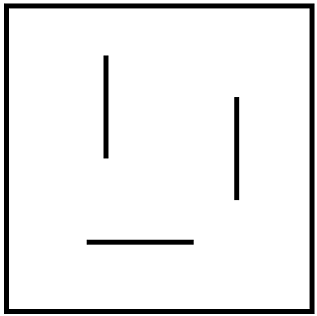}}
       \subfigure[]{\includegraphics[width=0.08\textwidth]{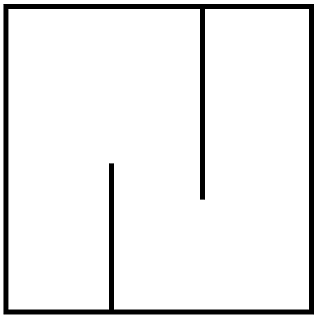}}
       \subfigure[]{\includegraphics[width=0.08\textwidth]{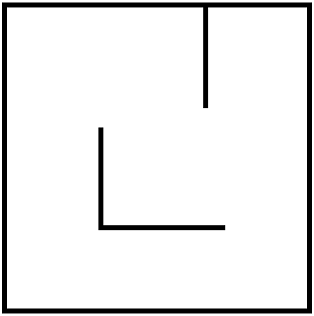}}
       
       \caption{10 testing maps. The size of each map is $6\times6m^2$, expressed in the coordinate system as (-3, -3) to (3, 3). The robot is supposed to start from (-2, -2) and reach the goal (2, 2). } 
       \label{fig:testmaps}
\end{figure}

Our network shares the same weights and parameters at each iteration step. It is comprised of two convolutional layers to convolve 3*180 dimensional data $\textbf{o}_d^t$ to 1*256, then it is concatenated with other two inputs $\textbf{o}_g^t$ and $Q_{i-1}$ to form 260 (256+2+2) dimensional data. Then it comes through a fully connected layer that is added with leaky rectified linear units (ReLUs). The output of the network are variables: mean and standard deviation of Gaussian Distribution, as required by the PPO method to generate continuous and more diverse action space, followed by a sampling method thus to return final positional point as our action space. After all iteration steps are finished, our final action space as path points are obtained by concatenating all single-step action space, shown in Equation \ref{eq:acts}.

\subsection{Training}
For training, we refer to multi-process training approaches (e.g., \cite{PPO, IJRR1, IJRR, PRIMAL}) to improve the training efficiency. Multi-process Proximal Policy Optimization (PPO)\cite{IJRR} enables a multi-process update of the network. Twenty-four robots are trained in the corresponding maps (shown in Figure \ref{fig:trainingmap}) and their start points and end points are set randomly. All policy training is done in simulation only. 

%When testing the policy on the simulation maps shown in Figure \ref{fig:testmaps}, it is clear that the gap will not exist. But if test the policy in real scenarios, it is obvious that the robot does not move as it does in the simulator. The benefits of fine-tuning can be seen at this point.}
% Assuming that there are M robots, then M robots generate actions based on the network at the same time, which makes the network update faster. Using this method to update the network has a huge advantage in computation time. We start 24 processes for training and we start 1 process when testing the models. The training map is shown in Figure \ref{fig:trainingmap}, and the testing map is shown in Figure \ref{fig:testmaps}. In order to reduce the complexity of path point generation, we specify that the distance $\rho$ between adjacent path points in the same path generation network model is constant and train only the angular deflections $\alpha_i$.

In order to find the best path representation, we trained a number of different path generation networks. We tested the path with different number of path points $N$ and with different distance $\rho$ of adjacent path points to find out the best-fit path composition. 

\subsection{Motion Fine-Tuning}

\begin{figure}[t]
       \centering
       \includegraphics[width=0.48\textwidth]{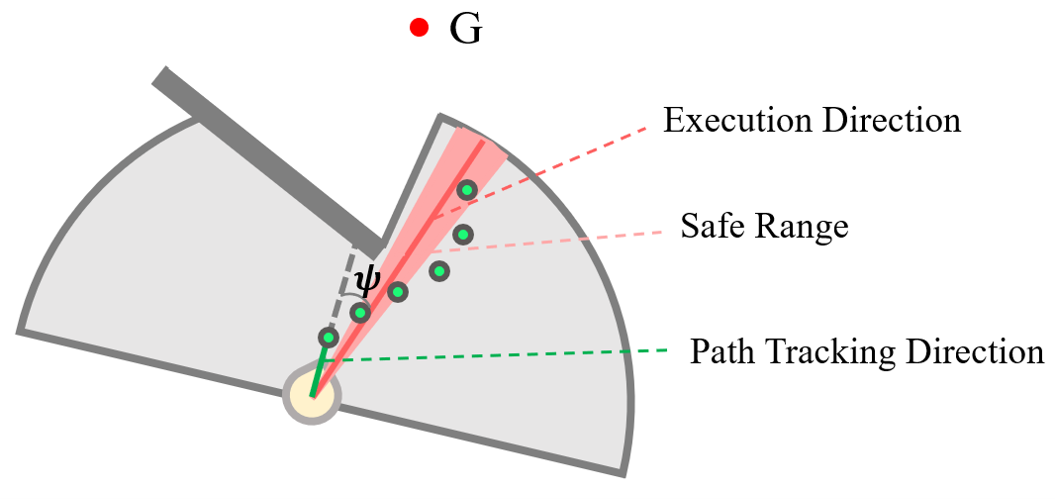}
       \caption{The motion fine-tuning module. Robot is supposed to follow the first predicted point generated by RL, which may lead to a collision if the direction is not safe. So the fine-tuning module is set to find a safe direction when generating the moving instructions for the robot. Execution Direction is selected by the minimum angle $\psi$ with the path tracking direction inside the Safe Range. The Safe Range here is a sector with an angle of 10 degree. }
       \label{fig:finetune}
\end{figure}

When the robot moves towards the target, the predicted points may be generated on both sides of robot at different time series, which may lead to shakes around predicted path. When the robot are too closed to an obstacle, if the shakes happens, the robot may crash into the obstacle.
Thus to ensure the safety of the robot, a module is provided to fine-tune the robot's motion instruction based on the predicted path points and the obstacle data, shown in Figure \ref{fig:finetune}. We calculate the minimum value $F(i)$ in Equation \ref{eq:F} to find the execution direction. Here, we set $\alpha = 0.3$ and $\beta = 1.0$. $i$ represents one of the 180 obstacle data, and the max\_range here represents the maximum sensor range and its value is 3.5m. $^3d$ is the 3rd latest obstacle data scanned by lidar, mentioned in Equation \ref{eq:ORD}. $\psi$ represents the angle between direction $i$ and the first path point, shown in Figure \ref{fig:finetune}.

\begin{equation}
    F(i) = \alpha \sum_{j=i-5}^{i+5} (^3d_j-max\_range)^2 + \beta\psi^2
    \label{eq:F}\tag{16}
\end{equation}
\begin{equation}
    i = \underset{i \in [-85, 85)\cap \mathbb{N}}{{\arg \min} \, F(i)}
    \label{eq:iF}\tag{17}
\end{equation}
% safe range 

We record $i$ when $F(i)$ is minimal and regard it as the execution direction. Here $i$ is in the range [-85, 85) to ensure the value $^3d$ exists in Equation \ref{eq:F}. Here we set the first point of the predicted path as the main moving direction. For the obstacle data scanned by lidar, if the obstacles are within 0.5m of the robot, the motion is fine-tuned to a direction away from the obstacles. 
% In our case, we selected a 10 degree sector for the safe range.

% When we generate a path by reinforcement learning, the challenge of getting the robot to issue motion commands using that path becomes a critical problem. In our approach, a simple obstacle avoidance algorithm (e.g., APF) is embedded in the controller. The obstacle avoidance algorithm needs to fine-tune the generated paths to improve safety performance. After the robot acquires the reference path, the obtained path is fine-tuned in combination with the obstacle avoidance algorithm and finally, the executable instructions of the robot are deduced.

% The controller can modify the robot's current driving direction if necessary. The path generation and execution process are asynchronous. If the robot does not execute the entire path completely for a certain interval and a new path is regenerated based on the latest position information of the robot, then the robot will follow the new reference path. At this point, the controller will discard the original path and regard the new path as the reference path to modify the current driving direction.

\section{Results and Discussion}

We demonstrate that our proposed approach has a better structure than other approaches through comparative experiments. Meanwhile, we aim to find the best path structure by ablation experiments.

In our comparison experiments, we focus on the comparison with the DWA-RL approach proposed in \cite{DWARL} and traditional approach Artificial potential field (APF) approach\cite{APF, APF1}. To compare the corresponding effects, the obstacle information acquisition of APF and DWA-RL approaches are adjusted accordingly in the experiment environments.

In the ablation experiments, we test the performance of different number of path points $N$ and the distance $\rho$ between adjacent path points.
% During the experiments, we found that the more path points, the longer the required network back-propagation process and the longer the training time. More iterations are needed if a better network model is to be trained. However, more path points and more complex paths better represent the real situation of the paths.

% \begin{table*}[!htbp]
% \begin{center}
% \caption{ \textsc{The Performance of Different Numbers of Points}}
% \begin{tabular}{|c|c|c|c|c|}
% \hline
% total length = 1m & Uncollided Paths & Sum Paths & Uncollided Rate (\%)& Success Rate (\%) &
% \cline{1-5}
% 3 points & 1057.40 & 1065.10 & 99.28 & 66.67 &
% \cline{1-5}
% 5 points & 693.33 & 1279.00 & 90.35 & 93.33 &
% \cline{1-5}
% 10 points & 548.27 & 588.47 & 93.17 & 100.00 & 
% \hline
% \end{tabular}
% \end{center}
% \label{table:PNum}
% \end{table*}
\subsection{Evaluation Metrics}
The following metrics are used to compare the performance of our approach with other approaches and to make further analysis in ablation studies. 
\begin{itemize}
    \item \textbf{Average Trajectory Length} - Average length of trajectory the robot travels from start point to goal point in the same test map.
    \item \textbf{Time Cost} - The average time cost from start point to goal point in the same test map.
    \item \textbf{Success Rate} - The rate of robot successfully travelling in an episode without any collision and finally reaching the goal. If a collision happens, the current test is immediately terminated and marked as failed.
    
    %\item \textbf{Success Rate of Predictive Paths} - The proportion of predictive paths that do not collide with obstacles and point to the goal point to the total predictive path the robot generated in an episode.
\end{itemize}
\subsection{Comparison Experiments}
\subsubsection{Simulation Experiments} 

\begin{figure}[htbp]
       {\includegraphics[width=0.48\textwidth]{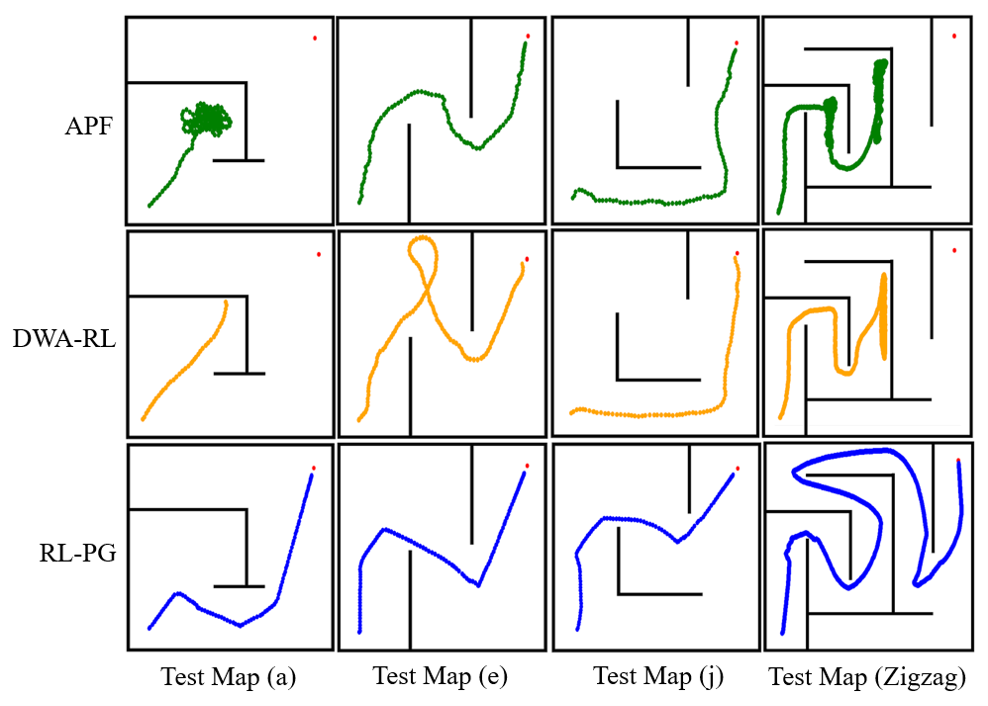}}
       
       \caption{We compare our model with DWA-RL approach and APF approach in testing maps shown in Figure \ref{fig:testmaps}, and we take map (a), (e), (j) and a Zigzag test map to show the real trajectory driven by the robot. The red points represent the goal point in different maps. The green trajectories represent the trajectories drawn by robot with APF algorithm, the orange trajectories represent the trajectories drawn by robot with DWA-RL approach, and the blue trajectories represent the trajectories drawn by robot with our RL-based Path Generation (RL-PG) approach.} 
       \label{fig:E2Ecom}
\end{figure}

%% textbf
\begin{table*}[!htbp]
    \centering
    \begin{tabular}{cccccccccccc}
    \toprule
    Metrics & Methods & Map a &  Map b & Map c & Map d & Map e & Map f & Map g & Map h & Map i & Map j \\
    \midrule
    \multirow{3}*{\textbf{Average Trajectory Length (m)}} 
    & APF & $\infty$ & \textbf{6.071} & 12.008 & \textbf{5.328} & 11.223 & 6.419 & 5.444 & 5.781 & 22.027 & 7.372\\
    & DWA-RL & $\infty$ & 6.178 & $\infty$ & 5.386 & 11.419 & 8.470 & \textbf{5.430} & 7.206 & 12.619 & \textbf{6.901}\\
    & RL-PG & \textbf{9.784} & 6.097 & \textbf{5.969} & 5.385 & \textbf{8.330} & \textbf{5.985} & 5.433 & \textbf{5.550} & \textbf{11.070} & 7.250\\
    \midrule
    \multirow{3}*{\textbf{Average Time Cost ($\times$10s)}} 
    & APF & $\infty$ & 6.620 & 12.508 & 5.971 & 11.658 & 7.066 & 6.156 & 6.402 & 22.340 & 7.862\\
    & DWA-RL & $\infty$ & 6.681 & $\infty$ & 5.938 & 12.093 & 8.697 & \textbf{5.938} & 7.561 & 13.187 & \textbf{7.231}\\
    & RL-PG & \textbf{10.207} & \textbf{6.607} & \textbf{6.496} & \textbf{5.728} & \textbf{8.937} & \textbf{6.210} & 5.944 & \textbf{6.044} & \textbf{11.706} & 8.038\\
    \midrule
    \multirow{3}*{\textbf{Success Rate (\%)}} 
    & APF & 0 & 100 & \textbf{100} & 100 & 100 & 100 & 100 & 100 & 70 & 100\\
    & DWA-RL & 0 & 100 & 0 & 100 & 100 & 100 & 100 & 100 & 20 & 100\\
    & RL-PG & \textbf{100} & 100 & \textbf{100} & 100 & 100 & 100 & 100 & 100 & \textbf{100} & 100\\
    \bottomrule
    \end{tabular}
    \caption{Comparison Experiment Results}
    \label{tab:compare_ete}
\end{table*}
\begin{figure}[htbp]
       \centering
       \subfigure[]{\includegraphics[width=0.32\textwidth,height=0.14\textwidth]{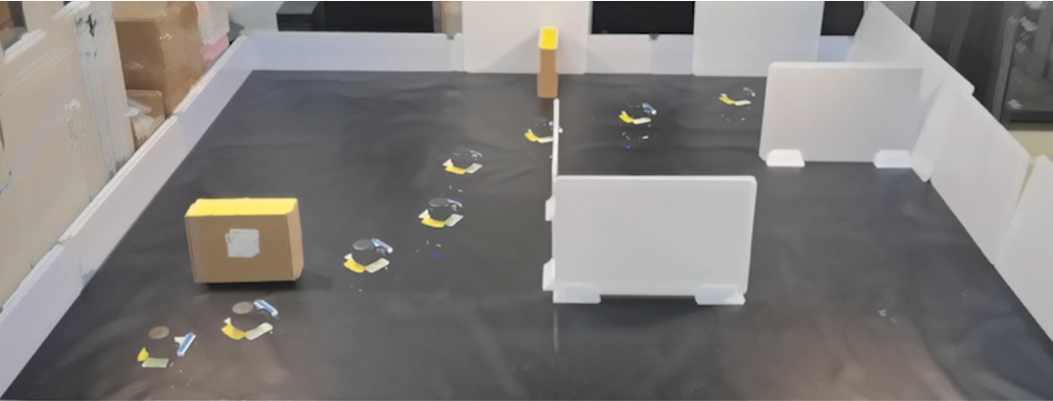}}
       \subfigure[]{\includegraphics[width=0.15\textwidth,height=0.14\textwidth]{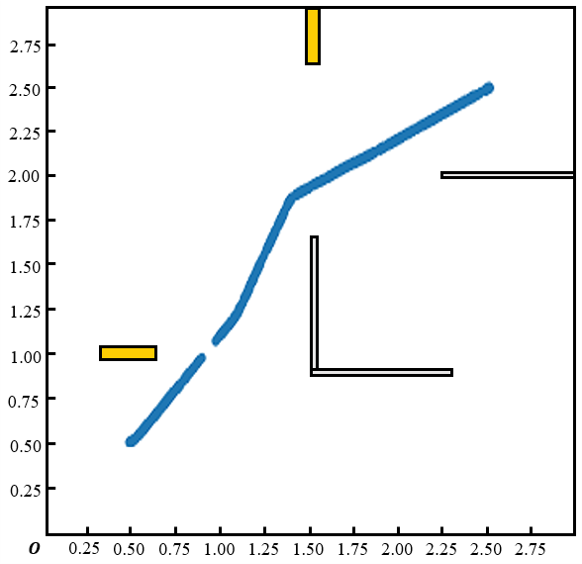}}
       \subfigure[]{\includegraphics[width=0.32\textwidth,height=0.14\textwidth]{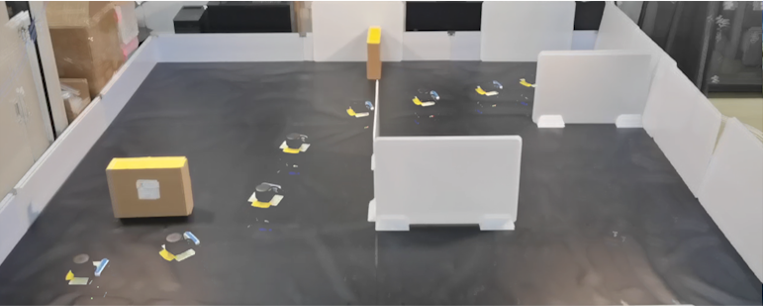}}
       \subfigure[]{\includegraphics[width=0.15\textwidth,height=0.14\textwidth]{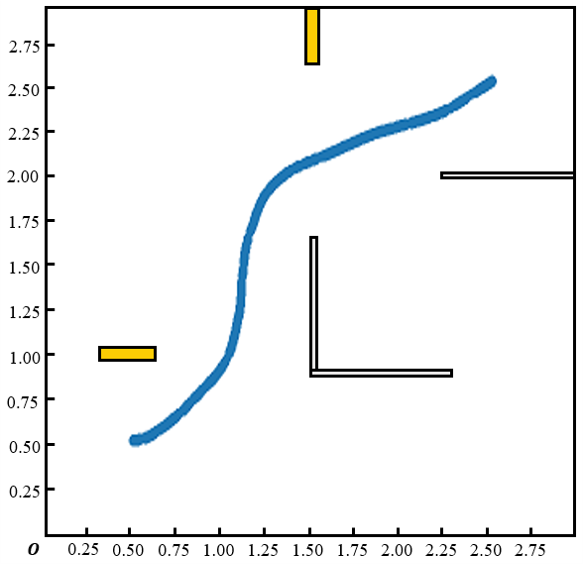}}
       \subfigure[]{\includegraphics[width=0.32\textwidth,height=0.14\textwidth]{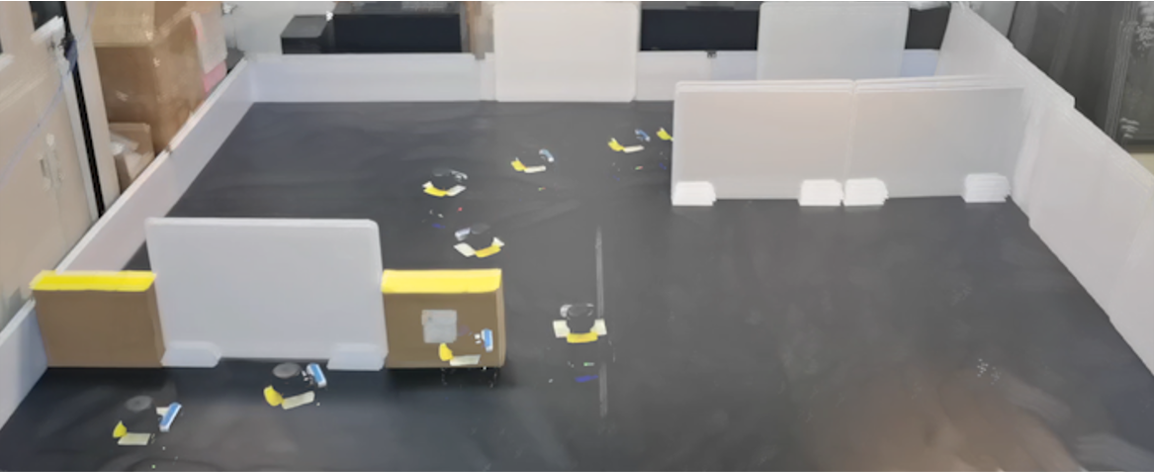}}
       \subfigure[]{\includegraphics[width=0.15\textwidth,height=0.14\textwidth]{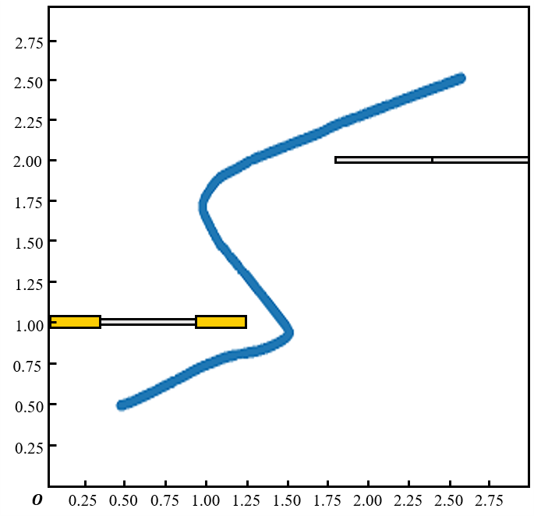}}
       \subfigure[]{\includegraphics[width=0.32\textwidth,height=0.14\textwidth]{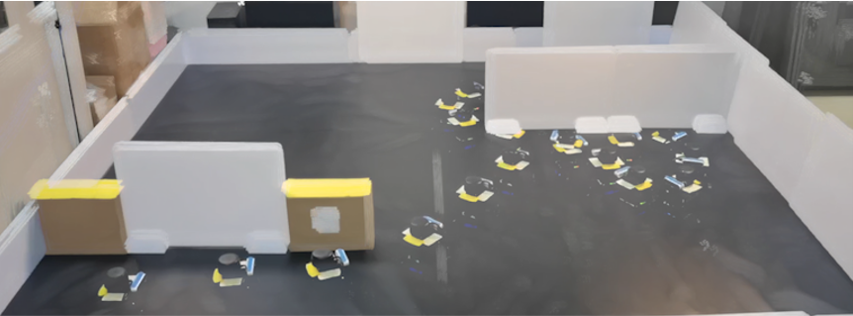}}
       \subfigure[]{\includegraphics[width=0.15\textwidth,height=0.14\textwidth]{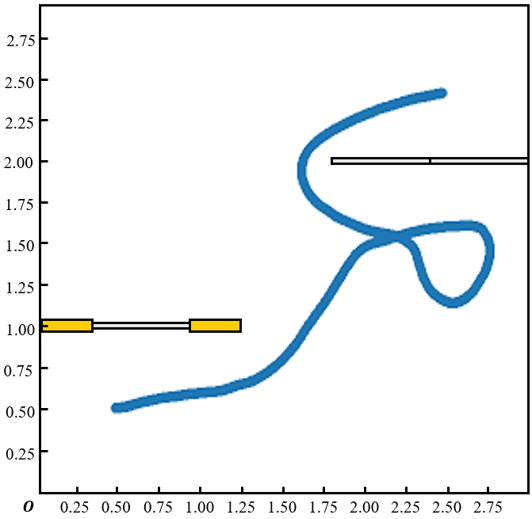}}
       
       \caption{(a) and (e) are the real scenario screenshots of RL-PG approach while (b) and (f) are their corresponding trajectories. (c) and (g) are the real scenario screenshots of DWA-RL approach while (d) and (h) are their corresponding trajectories.} 
       \label{fig:realpic}
\end{figure}

\begin{table*}[!htbp]
    \centering
    \begin{tabular}{cccccccc}
    \toprule
    \multirow{2}*{Metrics} & \multirow{2}*{~~~~$N$~~~~} & \multicolumn{6}{c}{$\rho$ (m)}\\
    \cmidrule{3-8}
    && ~~0.05~~ & ~~0.10~~ & ~~0.15~~ & ~~0.20~~ & ~~0.25~~ & ~~0.30~~ \\
    \midrule
    \multirow{4}*{\textbf{Average Trajectory 
    Length (m)}}
    & 3 & 5.93 & 6.02 & 6.00 & 6.04 & 5.92 & 5.90\\
    & 5 & 5.75 & 5.86 & 5.86 & 5.79 & 5.88 & 5.80\\
    & 10 & 5.65 & 5.69 & 5.70 & 5.68 & 5.66 & \textbf{5.63}\\
    & 15 & 5.77 & 5.77 & 5.80 & 5.81 & 5.79 & 5.75\\
    
    \midrule
    \multirow{4}*{\textbf{ Average Time 
    Cost ($\times$10s) }}
    & 3 & 8.82 & 8.82 & 9.38 & 9.07 & 9.28 & 8.89\\
    & 5 & \textbf{7.45} & 7.68 & 7.60 & 7.48 & 7.78 & 7.60\\
    & 10 & 7.57 & 7.67 & 7.71 & 7.50 & 7.63 & 7.51\\
    & 15 & 10.93 & 11.84 & 11.76 & 12.24 & 11.52 & 11.69\\
    
    \midrule
    \multirow{4}*{\textbf{ Success 
    Rate ($\%$) }}
    & 3 & 100 & 100 & 100 & 100 & 100 & 100\\
    & 5 & 100 & 100 & 100 & 100 & 100 & 100\\
    & 10 & 100 & 100 & 100 & 100 & 100 & 100\\
    & 15 & 100 & 80 & 100 & 100 & 90 & 90\\
    
    \bottomrule
    \end{tabular}
    \caption{Ablation Experiment Results}
    \label{tab:ablation}
\end{table*}

We test our approach with DWA-RL approach\cite{DWARL} and APF approach\cite{APF, APF1} on the same 10 testing maps shown in Figure \ref{fig:testmaps} based on the simulator \textbf{Stage}. We trained our policy on a computer with one i7-8700 CPU and one Nvidia GTX 2080s GPU. The training of both our approach and the DWA-RL approach\cite{DWARL} runs $120000$ episodes. We take the testing maps (a)(e)(j) in Figure \ref{fig:testmaps} and a Zigzag testing map as examples to show the trajectories robot travelled. Compared with other approaches, the trajectories of our approach are shorter, shown in Figure \ref{fig:E2Ecom}. We also found that when encountering certain situations, such as maze-like maps, the running time of the robot with DWA-RL and APF approaches increases significantly. But the robot with the path generation approach immediately finds a way to escape the maze and remove the deadlock situation caused by the environment. 
%% add a pic of apf and dwarl
%% different color to draw the trajectory

In Zigzag testing map shown in Figure \ref{fig:E2Ecom}, we can see that the robot of the other approaches cannot find a feasible path over the wall when encountering corners and long walls, shown in Test Map (Zigzag) in Figure \ref{fig:E2Ecom}. Our approach can make better choices in the direction close to the goal point by exploring feasible paths in the same case. The DWA-RL approach and APF approach consumes a lot of resources to explore other feasible solutions with lower success rate when faced with the limitations of the complex environment. 

Table \ref{tab:compare_ete} shows the detailed comparison results of the robot navigation tasks with different approaches. In test map (a), DWA-RL and APF approaches failed to find the goal due to the deceptive obstacles. In testing map (c) and (i), DWA-RL or APF also did not perform well. For trajectory length, our RL-PG approach are better in a deceptive map shown in Figure \ref{fig:E2Ecom}, while DWA-RL or APF approaches perform better in a simple environment. The max speed are all set 0.1m/s and we can find the RL-PG cost less time in most testing maps. As for the success rate, our RL-PG are all 100\%, while DWA-RL and APF failed in some of the testing environments. 
%% the table analyse

% Figure \ref{fig:totalstepssim} shows the number of neural network inferences for our approach and the End-to-End approach\cite{IJRR} during the robot's movement from the start point to the goal point. This demonstrates the greater impact of the environment on the End-to-End approach, requiring the neural network to make more inferences and consume more resources during the completion of the movement task.
% \begin{figure}[t]
%        \centering
%        \includegraphics[width=0.4\textwidth,height=0.2\textwidth]{figures/table5.png}
%        \caption{In simulation scenarios (testing maps (a)-(j) shown in Figure \ref{fig:testmaps}), the comparison of the networks in two approaches on the Number of Inferences. The blue bars and the red bars represent the Number of Inferences performed by the End-to-End approach and our approach respectively. Note that the value of the blue bar in Scenario (a) is out of range, which means the robot with the End-to-End approach failed to reach the goal in testing map (a).}
%        \label{fig:totalstepssim}
% \end{figure}
% \begin{figure}[t]
%        \centering
%        \includegraphics[width=0.35\textwidth,height=0.2\textwidth]{figures/table4.png}
%        \caption{In real scenarios, the comparison of the networks in two approaches on the Number of Inferences. The green bars and the yellow bars represent the Number of Inferences performed by the End-to-End approach and our approach respectively.}
%        \label{fig:totalsteps}
% \end{figure}
\subsubsection{Real Scenario Experiments}
We designed physical experiments using Turtlebot3 robot with RPlidar\_A2 on a $3\times3m^2$ area and the data is communicated through ROS. We reduce the map size compared to the scene of the simulated environment. 

% We found that the Number of Inferences calculated by our approach is less, as shown in Figure \ref{fig:totalsteps}. 

Some experimental screenshots and their corresponding trajectories are shown in Figure \ref{fig:realpic}. We found the pitfalls of the DWA-RL approach in real-world settings. In the real experimental scene, the DWA-RL approach trajectories are longer and the time costs are more than our approach as shown in Figure \ref{fig:realpic}. 
\subsection{Ablation Studies}
In the simulation environment, we did ablation studies to observe the effects by the different distances $\rho$ between adjacent path points and different number of the path points $N$. 
We trained each model for 1000 episodes to compare the results. Here we recorded the test data shown in Table \ref{tab:ablation}. 
We take testing map (h) in Figure \ref{fig:testmaps} as the test environment.
From an overall perspective, when $N=10$ and $\rho=0.30$, the average trajectory length is minimum, and when $N=5$ and $\rho=0.05$, the average time cost is minimum. When $N$ is large, the success rate is lower. 

We find that the performances of different distances between adjacent path points are close. The difference is that with smaller distances between adjacent path points, the network infers a higher probability of a predicted path that does not collide with an obstacle. The reason is that the larger the $\rho$, the easier the generated path is to conflict with the scanned obstacles.
For $N$, when $N$ is small, the path points are more dispersed and require greater angular deflection in path tracking, which makes robot consume more time to turn a larger angle. However when $N$ is large, the time cost also increases and it takes more time to find the feasible, uncollide paths.

% \subsubsection{If Fine-tune}

% We unload the fine-tuning module off the robot respectively to compare the effects. We used the same ten scenarios and the results are shown in Figure \ref{fig:PIDVFH}. 

% Here, about 60\% of the trajectories without the fine-tuning module reached the goal while 100\% of the robot with fine-tuning module reached the goal. 

% For the smoothness of the trajectories, shown in Figure \ref{fig:PIDVFH}, we found that the smoothness of the trajectories without the fine-tuning module was slightly better than that with the fine-tuning module. The reason is that the fine-tuning module needs to consider the angular velocity when the robot gets too close to the obstacle, which leads to an increase in the angle at which the actual path deviates from the originally generated path. From this point of view, the fine-tuning module improves the success rate of the robot to complete the task by losing the smoothness to a certain extent.

\section{Conclusion}

In this paper, we propose a novel RL-based robot path generation (RL-PG) approach with fine-tuned motion control. A deep Markov model suitable for path generation is trained using RL approach, which generates local paths in complex environments.
The corrective effect of the controller is also able to correct the path generated by the deep Markov model to avoid collisions and improve the safety performance of the robot system.

With the experiments of comparing our approach with DWA-RL approach and traditional APF approach, we demonstrate that the RL-based path generation (RL-PG) with fine-tuned motion control is more effective and more safe. What is more, our approach is able to find feasible paths in a complex maze-like environment. In future work, we will try to apply the approach of path generation in the motion planning for multiple robots.

% \bibliographystyle{IEEEtran}
% \bibliography{ref.bib}

\end{document}